\documentclass[letterpaper, 10 pt, journal, twoside]{ieeetran}
\usepackage{enumitem}
\usepackage{hyperref}
\usepackage{bm}
\usepackage{color}
\usepackage{cite}
\usepackage{amsmath,amssymb,amsfonts,amsthm}
\usepackage{algorithmic}
\usepackage{graphicx}
\usepackage{textcomp}

\newtheorem{thm}{Theorem}

\newtheorem{cor}[thm]{Corollary}

\theoremstyle{definition}
\newtheorem{defn}{Definition}[section]

 \newcommand{\rev}[1]{#1}

\begin{document}

\title{Robustness against Adversarial Attacks in Neural Networks using Incremental Dissipativity}
%
%
%

\author{Bernardo~Aquino, 
        Arash~Rahnama, 
        Peter~Seiler, 
        Lizhen~Lin,
        and~Vijay~Gupta 
\thanks{B. Aquino and V. Gupta are with the Department of Electrical Engineering, University of Notre Dame, Notre Dame,
IN, 46656 USA. Email: {\tt{\{bcruz2,vgupta\}@nd.edu}} A. Rahnama is with Amazon Inc. Email: {\tt{arashrahnama@gmail.com}} P. Seiler is with the Electrical Engineering and Computer Science, University of Michigan, Ann Arbor, MI 48109, USA. Email: {\tt{pseiler@umich.edu}} L. Lin is with the Department of Applied Computational Mathematics and Statistics, University of Notre Dame, Notre Dame,
IN, 46656 USA. Email: {\tt{lizhen.lin@nd.edu}}}
}

\maketitle
\thispagestyle{empty}
\begin{abstract}
Adversarial examples can easily degrade the classification performance in neural networks. Empirical methods for promoting robustness to such examples have been proposed, but often lack both analytical insights and formal guarantees. Recently, some robustness certificates have appeared in the literature based on system theoretic notions. This work proposes an incremental dissipativity-based robustness certificate for neural networks in the form of a linear matrix inequality for each layer. We also propose  a sufficient spectral norm bound for this certificate which is scalable to neural networks with multiple layers. We demonstrate the improved performance against adversarial attacks on a feed-forward neural network trained on MNIST and an Alexnet trained using CIFAR-10.
\end{abstract}

\begin{IEEEkeywords}
Adversarial Attacks, Deep Neural Networks, Robust Design, Passivity Theory, Spectral Regularization
\end{IEEEkeywords}

%

\section{Introduction}
%
%
%
%
\IEEEPARstart{N}{eural} networks are powerful structures that can represent any non-linear function through appropriate training for classification and regression tasks. However, neural networks still lack formal performance guarantees, limiting their application in safety-critical systems \cite{szegedy2014intriguing}. Furthermore, many studies have shown the susceptibility of neural networks to small perturbations through carefully crafted adversarial attacks, which, in the case of imaging systems, may be imperceptible to the human eye \cite{szegedy2014intriguing}, \cite{carlini2017adversarial}, \cite{athalye2018obfuscated}, \cite{su2019one}. 

Different types of defenses have  emerged trying to address this shortcoming, with perhaps the most successful of them being adversarial training \cite{ganin2016domain}, \cite{shafahi2019adversarial}, \cite{tramer2017ensemble}, \cite{robey2021adversarial} and defensive distillation \cite{papernot2016distillation}, \cite{papernot2016effectiveness}. However, even though neural networks with these defenses empirically show superior performance against adversarial attacks than those without such approaches, these methods do not broadly provide either design insights or formal guarantees on robustness.

In the search for guarantees, other studies have proposed systems-theoretic robustness certificates during training. For instance, Lyapunov stability certificates with incremental quadratic constraints were proposed in ~\cite{arcakstability} and imitation learning in \cite{yin2021imitation}. Lipschitz constant-based certificates have been a particularly fruitful approach, of which we highlight  \cite{fazlyab2019efficient}, which uses a convex optimization approach, \cite{latorre2020lipschitz} with sparse polynomial optimization, \cite{combettes2020lipschitz}, which uses averaged activation operators, and \cite{pauli2021training}, which proposes global Lipschitz training using the Alternate Direction Method of Multipliers (ADMM) method. 

In this work, we take an incremental dissipativity-based approach and derive a robustness certificate for neural network architectures in the form of a Linear Matrix Inequality (LMI). This LMI utilizes Incremental Quadratic Constraints (iQCs) to describe non-linearities in each neuron as proposed in~\cite{fazlyab2020safety}. Recently,~\cite{revay2021recurrent} used iQCs to define a new class of Recurrent (Equilibrium) Neural Networks for applications such as system identification and robust feedback control. We derive a sufficient condition from this LMI as a bound on the spectral norm of the weight matrix in each layer that is easily implementable. There are three advantages of this approach. One, this approach generalizes the Lipschitz condition that has been imposed on neural networks in previous works to a condition guaranteeing that the input output response of the neural network is sector bounded. The Lipschitz condition defines a particular sector that can be expressed as a special case in our framework. Providing more degrees of freedom to the designer in choosing the sector can lead to a better point in the performance-robustness tradeoff.  Two, this approach can scale with the number of layers of the neural network. Consideration of deep neural structures and convolutional layers is a known issue with the existing optimization based paradigms~\cite{abadi2016tensorflow} and has been explicitly pointed out as a limitation in works such as~\cite{pauli2021training}. Three, our condition provides an insight on the high empirical effectiveness of the usage of spectral norm regularization for improved generalizability in deep network structures \cite{yoshida2017spectral} and training stability of Generative Adversarial Networks (GANs) \cite{miyato2018spectral}.  


Passivity analysis for neural networks has been considered  for systems with time delay \cite{li2005passivity}, \cite{xu2009passivity}, but no training methodology was offered, and the resulting LMIs can quickly become computationally burdensome.~\cite{Rahnama_2020_CVPR} provided a passivity approach for robust neural networks, but no certificate was presented. Further, our theoretical development is for the more general notion of incremental sector bounded neural network layers, which allows for negative passivity indices.


\section{Background}\label{formulation}


In this work, we propose an incremental dissipativity-based approach to quantify and engineer the robustness of neural networks against adversarial attacks. We focus on image classification networks, since the design of adversarial perturbations is perhaps the most developed in such systems~\cite{goodfellow2014explaining,madry2017towards}. Further, the presence of convolutional layers in most of such architectures poses  challenges to existing approaches for guaranteeing robustness. In such applications, our main goal is to reduce the classification error rate on adversarial image test sets. One should notice that the proposed approach can be applied to other applications as well. 

\paragraph*{Robustness against adversarial attacks} Adversarial attacks on neural networks seek to produce a significant change in the output when the input is perturbed slightly. Thus, designing a network that limits the change in the output as a function of the change in the input can mitigate the effects of adversarial attacks. 
%
%
We claim that enforcing neural network systems to be incrementally dissipative (specifically, sector bounded) can limit the variation on the output, given some perturbation in the input. 

Consider a neural network $f_{NN}: \mathcal{R}^n \to \mathcal{R}^m$. In the image classification example, the input may be the vectorized version of the image and the output may be a vector of probabilities corresponding to the various classes. We are interested in the deviation between $f_{NN}(x)$ and $f_{NN}(x+\delta_x)$ where $x$ and $x+\delta_x$ are the actual and adversarial inputs, respectively. Recently, a number of works have pointed out that by enforcing the condition 
\begin{equation}
\label{eq:lip}
    \left\Vert f_{NN}(x+\delta_x)-f_{NN}(x)\right\Vert_2\leq \gamma\left\Vert\delta_x\right\Vert_2
\end{equation}
for some $\gamma>0$, we can guarantee 
that a norm-bounded adversarial perturbation can shift the output of the neural network only by a bounded amount. Intuitively, this leads to a certain robustness in the classification performance of the neural network.  
While the Lipschitz constant based approach enforces a symmetric sector in the sense of equation~\ref{eq:lip}) above, the incremental sector boundedness constraint that we consider in this paper generalizes it by allowing the sector slopes to be independent of each other and have arbitrary signs. Further, we can consider convolutional layers as well as compose metrics for each layer to obtain a robustness certificate for the entire network. Finally, we can obtain a computationally efficient training approach to ensure that this certificate is met using a spectral norm based condition. In this sense, our work also sheds light on the empirically observed effectiveness of spectral norm based regularization to promote robustness in neural networks. However, this approach cannot guarantee robustness for all attacks and data sets. As stated in \cite{nar2019cross}, depending on the specific data set, the margin between decision boundaries can be very narrow, and even a tiny perturbation on output can change the classification label. Therefore, our method implicitly assumes there is some margin between classes.



\paragraph*{Sector boundedness} Dissipativity has been widely used in control systems theory due to its close relation to stability analysis and compositional properties \cite{passivebook}\cite{passtypes}\cite{passivebook2}.
%
\begin{defn}
(\cite{passdef}) A discrete-time system with input $\mathbf{x}(k)$ and corresponding output $\mathbf{y}(k)$ at time $k$ is $(Q,S,R)$ dissipative 
if $\forall k$, the condition 
$    0 \leq s(\mathbf{x}(k),\mathbf{y}(k))$ holds for all admissible initial conditions of the system and all admissible sequences of the inputs $\{\mathbf{x}(j)\}_{j=0}^{k},$
where $s(\mathbf{x}(k),\mathbf{y}(k))$, is the supply rate given by
\begin{equation}\label{supply}
    s(\mathbf{x},\mathbf{y})=\mathbf{y}^T\mathbf{Q}\mathbf{y}+\mathbf{x}^T\mathbf{S}\mathbf{y}+\mathbf{y}^T\mathbf{S}^T\mathbf{x}+\mathbf{x}^T\mathbf{R}\mathbf{x},
\end{equation}
for matrices $\mathbf{Q}$, $\mathbf{S}$, and $\mathbf{R}$ of appropriate dimensions.
\end{defn}
 Depending on the matrices $\mathbf{Q}$, $\mathbf{S}$, and $\mathbf{R}$, the system will exhibit different dynamical properties. We are particularly interested in the following three cases:
\begin{defn}
(\cite{passtypes},\cite{disstypes}) A $(Q,S,R)$ dissipative discrete-time system is:
\begin{enumerate}
  \item passive, if $\mathbf{Q}=\mathbf{0}$, $\mathbf{S}=\frac{1}{2}\mathbf{I}$, $\mathbf{R}=\mathbf{0}$.
  \item \label{passivecase}strictly passive, if $\mathbf{Q}=-\delta \mathbf{I}$, $\mathbf{S}=\frac{1}{2}\mathbf{I}$, $\mathbf{R}=-\nu \mathbf{I}$, for some $\delta>0$ and $\nu>0$.
  \item \label{case} sector bounded with slopes $ \frac{1-\sqrt{1-4\delta\nu}}{2\delta}$ and $\frac{1+\sqrt{1-4\delta\nu}}{2\delta}$, if $\mathbf{Q}=-\delta \mathbf{I}$, $\mathbf{S}=\frac{1}{2}\mathbf{I}$,$\mathbf{R}=-\nu \mathbf{I}$, for some $\delta$ and $\nu$. 
\end{enumerate}    
\end{defn}
For sector bounded systems, the constants $\delta$ and $\nu$ are called passivity indexes. 
We will call a sector bounded system output strictly passive (OSP) when $\delta>0$ and input strictly passive (ISP) when $\nu>0$. 
Finally, all the above definitions can be extended to the case of incrementally $QSR$-dissipative systems, incrementally passive systems, incrementally strictly passive systems, and incrementally sector bounded systems by defining the supply rate as 
\begin{multline}\label{supplydiff}
    s(\mathbf{\Delta_x},\mathbf{\Delta_y})=\mathbf{\Delta_y}^T\mathbf{Q}\mathbf{\Delta_y}\\+\mathbf{\Delta_x}^T\mathbf{S}\mathbf{\Delta_y}+\mathbf{\Delta_y}^T\mathbf{S}^T\mathbf{\Delta_x}+\mathbf{\Delta_x}^T\mathbf{R}\mathbf{\Delta_x},
\end{multline}
where $\mathbf{\Delta_x} = \mathbf{x_{1}}-\mathbf{x_{2}}$ and $\mathbf{\Delta_y} = \mathbf{y_{1}}-\mathbf{y_{2}}$ for two inputs $\mathbf{x_{1}}$ and $\mathbf{x_{2}}$, and the corresponding outputs $\mathbf{y_{1}}$ and $\mathbf{y_{2}}$.

\section{Enforcing Incremental Sector Boundedness}
\label{sec:enforcing}


\paragraph{Incremental QSR-dissipativity for a single layer}

Consider a neural network layer that receives the  vector $\mathbf{x}$ as an input and yields the output $\mathbf{y}$. For a non-convolutional layer, $\mathbf{y} = \phi(\mathbf{W}\mathbf{x}+\mathbf{b})$, where $\phi(\cdot)$ is an element-wise nonlinear activation function. In this paper, we assume that the function is incrementally sector bounded by $[\alpha ,\beta ]$ for some $\alpha$ and $\beta$. This assumption is satisfied by most commonly used functions including tanh, ReLU, and leaky ReLU functions, and is commonly made in works such as \cite{arcakstability}. We also assume that the same activation function $\phi(.)$ is used in every element of the layer, which is the case for most neural network layer architectures. 
We can then state the following result.

\begin{thm}
\label{theorem:single_layer}
    Consider a non-convolutional neural network layer defined by $\mathbf{y} = \phi(\mathbf{Wx+b})$ where $\phi(.)$ is incrementally sector bounded by $[\alpha,\beta]$. Define $m=\frac{\alpha+\beta}{2}$ and $p=\alpha\beta$. The layer is incrementally QSR dissipative if the LMI
   \begin{equation}\label{sprocineqstate}
        \mathbf{M} = \left[
        \begin{matrix}
        \mathbf{Q} & \mathbf{S} \\
        \mathbf{S}^\top & \mathbf{R}
    \end{matrix}\right]+\left[
        \begin{matrix}
        \mathbf{\Lambda} & -m\mathbf{\Lambda}\mathbf{W} \\
        -m\mathbf{W}^\top \mathbf{\Lambda} & p\mathbf{W}^\top \mathbf{\Lambda}\mathbf{W}
    \end{matrix}\right] \succcurlyeq 0
\end{equation}
    is feasible for some $\mathbf{\Lambda}$ defined as
    
    \begin{equation}\label{lambdadef}
    \mathbf{\Lambda}=\sum_{1\leq i \leq n}\lambda_{ii}\mathbf{e}_i\mathbf{e}_i^\top\geq0,
    \end{equation}
    where $\mathbf{e}_{i}$ is the $i$-th standard basis vector.
\end{thm}

\begin{proof}
The proof follows by using an S-procedure following the arguments in~\cite{fazlyab2020safety}. First, for notational ease, for any input $\mathbf{x_{i}}$ of the neural network layer, denote the output by $\mathbf{y_{i}}$ and define $\mathbf{v_{i}} = \mathbf{Wx_{i}}+\mathbf{b}$. With this notation, we can write:
\begin{equation}\label{equivalence}
    \left[\begin{matrix} \mathbf{0} & \mathbf{W}\\ \mathbf{I} & \mathbf{0}  \end{matrix}\right] \left[\begin{matrix} \mathbf{y_{1}}-\mathbf{y_{2}} \\ \mathbf{x_{1}}-\mathbf{x_{2}} \end{matrix}\right] =\left[\begin{matrix} \mathbf{v_{1}}-\mathbf{v_{2}}\\ \mathbf{y_{1}}-\mathbf{y_{2}} \end{matrix}\right],
\end{equation}
for any inputs $\mathbf{x}_{1}$ and $\mathbf{x}_{2}$. 

The first quadratic form for the S-procedure can be obtained using the iQC approach outlined in~\cite{fazlyab2020safety}. Since the nonlinear function $\phi(\cdot)$ is an element wise function that is incrementally sector bounded by $[\alpha ,\beta ],$ we have 
\begin{equation}\label{ineqsector3}
    \alpha  \leq \frac{\mathbf{y}_{1}^{j}-\mathbf{y}_{2}^{j}}{\mathbf{v}_{1}^{j}-\mathbf{v}_{2}^{j}} \leq \beta,
\end{equation}
for all $j=1,\ldots, n$, where $n$ is the number of neurons at the layer under consideration, $\mathbf{v}_{1}^{j}$ denotes the input to the nonlinear activation function for the $j$-th neuron, and $\mathbf{y}_{1}^{j}$ denotes its output. The collection of these equations can be written as
\begin{equation}\label{nonlinearineq3}
\left[\begin{array}{c}
  \mathbf{v_{1}-v_{2}} \\
  \mathbf{y_{1}-y_{2}} \\ 
\end{array}\right]^T
\left[\begin{array}{lr}
  p \mathbf{\Gamma} & -m\mathbf{\Gamma} \\ 
  -m\mathbf{\Gamma} & \mathbf{\Gamma}\\
\end{array}\right]
\left[\begin{array}{c}
  \mathbf{v_{1}-v_{2}} \\
  \mathbf{y_{1}-y_{2}} \\ 
\end{array}\right]\leq 0,
\end{equation}
for a matrix $\mathbf{\Gamma}$ of Lagrange multipliers with the structure defined in \eqref{lambdadef} and where $m = \frac{\alpha +\beta }{2}$ and $p=\alpha \beta$ \cite{fazlyab2019efficient}. Finally, \eqref{equivalence} yields:
\begin{equation}\label{nonlinearineqmod}
\left[\begin{array}{c}
  \mathbf{y_{1}-y_{2}} \\
  \mathbf{x_{1}-x_{2}} \\ 
\end{array}\right]^T
\left[\begin{array}{lr}
  \mathbf{\Gamma} & -m\mathbf{\Gamma}\mathbf{W} \\ 
  -m\mathbf{W}^{T}\mathbf{\Gamma} & p\mathbf{W}^{T}\mathbf{\Gamma}\mathbf{W}\\
\end{array}\right]
\left[\begin{array}{c}
  \mathbf{y_{1}-y_{2}} \\
  \mathbf{x_{1}-x_{2}} \\ 
\end{array}\right]\leq 0.
\end{equation}

The second quadratic form is obtained from the definition of incremental $(QSR)$ dissipativity as

\begin{equation}\label{qsrmatrixineq}
    \left[\begin{matrix} \mathbf{y}_{1}-\mathbf{y}_{2} \\ \mathbf{x}_{1}-\mathbf{x}_{2} \end{matrix}\right]^T\left[\begin{matrix} -\mathbf{Q} & -\mathbf{S}\\ -\mathbf{S}^T & -\mathbf{R} \end{matrix}\right] \left[\begin{matrix} \mathbf{y}_{1}-\mathbf{y}_{2}\\ \mathbf{x}_{1}-\mathbf{x}_{2} \end{matrix}\right] \leq 0.
\end{equation}
Thus, using an S-Procedure on \eqref{nonlinearineqmod} to enforce \eqref{qsrmatrixineq} we obtain:
\begin{equation}
\lambda\left[\begin{matrix}
\mathbf{\Gamma} & -m\mathbf{\Gamma}\mathbf{W} \\
-m\mathbf{W}^T\mathbf{\Gamma} & p\mathbf{W}^T \mathbf{\Gamma}\mathbf{W} \\
\end{matrix}\right]-\left[\begin{matrix} -\mathbf{Q} & -\mathbf{S}  \\ -\mathbf{S}^T & -\mathbf{R}\\
     \end{matrix}\right]
 \succcurlyeq 0,
\end{equation}
where $\lambda\geq 0$. The result now follows by defining $\mathbf{\Lambda}=\lambda\mathbf{\Gamma}.$
%
%
%
\end{proof}
For a convolutional layer, we can derive a similar result by using the result from \cite{sedghi2018singular} that the convolution operation with a filter is equivalent to a matrix multiplication of a block circulant matrix composed by the filter coefficients. Specifically, if the input image $\mathbf{X}$ is convolved with a filter with impulse response $\mathbf{F}\in\mathcal{R}^{n\times n}$, then define the doubly block circulant matrix $\mathbf{C}$ as the matrix
\begin{equation}
\label{eq:definition_C}
\left[\begin{matrix}
  \text{circ}(\mathbf{F(0,:)}) & \text{circ}(\mathbf{F(1,:)}) & \hdots & \text{circ}(\mathbf{F(n-1,:)}) \\ 
  \text{circ}(\mathbf{F(n-1,:)}) & \text{circ}(\mathbf{F(0,:)}) & \hdots & \text{circ}(\mathbf{F(n-2,:)}) \\ 
  \vdots & \vdots & \ddots & \vdots \\
    \text{circ}(\mathbf{F(1,:)}) & \text{circ}(\mathbf{F(2,:)}) & \hdots & \text{circ}(\mathbf{F(0,:)}) \\ 
\end{matrix}\right],
\end{equation}
where $\mathbf{F(p,:)}$ denotes the $p$-th row of the matrix $\mathbf{F}$ and $\text{circ}(\mathbf{v})$ with a vector $\mathbf{v}\in\mathcal{R}^{n\times 1}$ produces an $n\times n$ circulant matrix with the first row as the vector $\mathbf{v}$. Then,~\cite{Rahnama_2020_CVPR,sedghi2018singular} showed that the output $\mathbf{O}$ of the convolution can be expressed as
\begin{equation}
    \text{vec}(\mathbf{O}) = \mathbf{C}\text{vec}(\mathbf{X}),
\end{equation}
where $\text{vec}(.)$ is the standard vectorization operation. Thus, we can extend Theorem~\ref{theorem:single_layer} to a convolutional layer as follows.
\begin{cor}
    Consider the setting of Theorem~\ref{theorem:single_layer} but with a convolutional neural network layer in which the input is convolved with the filter $\mathbf{F}$, vectorized, and transmitted through an element-wise non-linear activation function $\phi(.)$ that is sector bounded by $[\alpha,\beta]$. Define $m=\frac{\alpha+\beta}{2}$ and $p=\alpha\beta$. The layer is incrementally QSR dissipative if the LMI
   \begin{equation}\label{sprocineqstate:2}
        \mathbf{M} = \left[
        \begin{matrix}
        \mathbf{Q} & \mathbf{S} \\
        \mathbf{S}^\top & \mathbf{R}
    \end{matrix}\right]+\left[
        \begin{matrix}
        \mathbf{\Lambda} & -m\mathbf{\Lambda}\mathbf{C} \\
        -m\mathbf{C}^{\top} \mathbf{\Lambda} & p\mathbf{C}^{\top} \mathbf{\Lambda}\mathbf{C}
    \end{matrix}\right] \succcurlyeq 0
\end{equation}
    is feasible for some $\mathbf{\Lambda}$ defined as
    
    \begin{equation}\label{lambdadef:2}
    \mathbf{\Lambda}=\sum_{1\leq i \leq n}\lambda_{ii}\mathbf{e}_i\mathbf{e}_i\geq0,
    \end{equation}
    where $\mathbf{e}_{i}$ is the $i$-th standard basis vector and $\mathbf{C}$ is defined as in~(\ref{eq:definition_C}).
\end{cor}

We note that by construction, the matrix $\mathbf{\Lambda}$ with the structure in~\eqref{lambdadef} is positive definite.





\paragraph{Extension for a multi layered 
neural network}
The argument given above can be extended to consider the entire neural network instead of one layer. However, as was noted in~\cite{pauli2021training} even for the LMIs resulting from the simpler Lipschitz constraint, this approach quickly becomes computationally cumbersome. Instead, we can utilize the compositional property of $(QSR)$-dissipativity to ensure that a multi-layered neural network is sector bounded by imposing constraints on each layer separately. \rev{We have the following result.}
\begin{thm}
\label{thm:entire_network}
   Consider a neural network with $n$ layers, where each layer $i$ is incrementally $QSR$ dissipative with $\mathbf{Q}=-\delta_{i}\mathbf{I}$, $\mathbf{S}=0.5\mathbf{I}$, and $\mathbf{R}=-\nu_{i}\mathbf{I}$. Then, the neural network is incrementally sector bounded with parameters $\mathbf{Q}=-\delta\mathbf{I}$, $\mathbf{S}=0.5\mathbf{I}$, and $\mathbf{R}=-\nu\mathbf{I}$ if the matrix $\mathbf{A}\preceq 0,$ where
   
    \begin{equation*}
        \mathbf{A} \triangleq \left[\begin{matrix}
        -\nu_1+\nu & \frac{1}{2} & 0 & \cdots & -\frac{1}{2}\\
        \frac{1}{2} & -\nu_2+\delta_1 & \frac{1}{2} & \cdots & 0 \\
        \vdots & \ddots & \ddots & \ddots & \vdots \\
        0 & \cdots & \frac{1}{2} & -\nu_n+\delta_{n-1} & \frac{1}{2} \\
        -\frac{1}{2} & 0 & \cdots & \frac{1}{2} & -\delta_n+\delta \\
        \end{matrix}\right].
    \end{equation*}  
\end{thm}
\begin{proof}
Proof follows directly be viewing the neural network as a cascade of layers and applying~\cite[Theorem 5]{cherry}.
\end{proof}

Theorem~\ref{thm:entire_network} thus provides one way of ensuring that the neural network is sector bounded, and hence, robust. Specifically, we can impose the constraint \eqref{sprocineqstate} during the training of the network. However, this requires solving an SDP problem on each gradient descent step or, at best, after a certain number of epochs, as described in \cite{pauli2021training}. While ensuring  $\mathbf{M}_i \succcurlyeq 0$, with $i=1,\ldots, l$ for each layer separately is computationally more tractable than optimizing the entire neural network through a large SDP, it nonetheless makes the training much slower. This makes it desirable to further reduce the computational complexity as discussed next. 

\section{Training Robust Neural Networks}\label{trainrobust}

In this section we derive a sufficient condition on the spectral norm of the weight matrix that  provides a feasible solution for the LMI \eqref{sprocineqstate} for sector bounded systems. We focus once again on a single neural network layer in the setting of Theorem~\ref{theorem:single_layer} for ease of notation. Denote the spectral norm of matrix $\mathbf{A}$ by $\Vert \mathbf{A} \Vert_2$. Further, define the infimum seminorm $\Vert \mathbf{A} \Vert_{i}$ as the square root of the minimum eigenvalue of $\mathbf{A}^\top\mathbf{A}$. While the spectral norm is sub-multiplicative, the infimum seminorm is super-multiplicative. Further, the two norms are related for invertible matrices through the relation  $\Vert\mathbf{A}^{-1}\Vert_2=\Vert \mathbf{A}\Vert_{i}^{-1}$~\cite[Equation 2.5]{feingold1962block}. Finally, for a normal matrix $\mathbf{\Gamma}$, we have~\cite[Theorem 2.5]{bennet79} $\|\mathbf{\Gamma}-I\|_{i}=\|\mathbf{\Gamma}\|_{i}-1.$

Note that for a sector bound on a neural network to lead to bounded perturbation of the output given a perturbation in the input, it is natural to assume that the slope of the upper bound on the sector given by $\delta$ is positive. The slope of the lower bound $\nu$ can have arbitrary sign, but should be bounded. 
We make these assumptions in presenting the following result. 

\begin{thm}\label{region}
Consider the setting of Theorem~\ref{theorem:single_layer} with $\mathbf{Q}=-\delta\mathbf{I}$, $\mathbf{S}=0.5\mathbf{I}$, and $\mathbf{R}=-\nu\mathbf{I}$ with $\delta>0$. If the following inequalities hold
\begin{align}
\label{eq:thm_cond_1}
     \left\Vert\mathbf{W}\right\Vert_2 &\leq  \frac{1}{\vert m\vert}\left(1-\frac{(\delta+0.5)(1- \vert p\vert\left\Vert \mathbf{W}\right\Vert_{i}^{2})}{\delta-\nu}\right)\\
     \vert p\vert\left\Vert\mathbf{W}\right\Vert_{i}^{2}&\leq 1, 
     \label{eq:thm_cond_2}
\end{align}
then (\ref{sprocineqstate}) is feasible with a matrix $\mathbf{\Lambda}$ of the form~(\ref{lambdadef}) with \begin{equation}
\label{eq:thm_cond_3}
\left\Vert\mathbf{\Lambda}\right\Vert_2=\left\Vert\mathbf{\Lambda}\right\Vert_{i}\geq\frac{\delta-\nu}{1-\vert p\vert\left\Vert\mathbf{W}\right\Vert_{i}^{2}}.
\end{equation}


\end{thm}

\begin{proof}
A sufficient condition for the feasibility of the LMI~(\ref{sprocineqstate}) is that the matrix $\mathbf{M}$ be block diagonally dominant with diagonal blocks positive semi-definite~\cite{feingold1962block}. In other words, if there exists a matrix $\mathbf{\Lambda}$ of the form~(\ref{lambdadef}) that ensures that the following four conditions are satisfied, then the LMI~(\ref{sprocineqstate}) is feasible for that $\mathbf{\Lambda}$:
\begin{align}
\mathbf{\Lambda+Q} &\succ 0\label{pd1}\\
\mathbf{W}^\top p\mathbf{\Lambda W+R} &\succ 0\label{pd2}\\
\left\Vert(\mathbf{Q}+\mathbf{\Lambda})^{-1}(\mathbf{S}-m\mathbf{\Lambda}\mathbf{W})\right\Vert_2 &\leq   1\label{sn1}\\
\left\Vert(\mathbf{W}^\top p\mathbf{\Lambda W+R})^{-1}(\mathbf{S}^\top-m\mathbf{W}^\top\mathbf{\Lambda})\right\Vert_2 &\leq   1\label{sn2}.
\end{align}
We now consider these conditions one by one.
\begin{enumerate}
\item {Claim 1: Any $\mathbf{\Lambda}$ that satisfies equation
\begin{equation}\label{Lmin}
    \Vert\mathbf{\Lambda}\Vert_{i}\geq \delta,
\end{equation}
will ensure that \eqref{pd1} is satisfied.} This claim follows by noting that  $Q=-\delta \mathbf{I}$, $\delta>0$, and $\mathbf{\Lambda}\succcurlyeq 0.$ 

\item 
Claim 2: Any $\mathbf{\Lambda}$ that satisfies the condition

\begin{equation}\label{boundclaim2}
    \vert p\vert\Vert\mathbf{W} \Vert_{i}^2\Vert\mathbf{\Lambda}\Vert_{i}
    > \nu
\end{equation}
will ensure that \eqref{pd2} is satisfied. This is because
\begin{align*}
    \eqref{pd2}&\Longleftrightarrow \vert p\vert\Vert\mathbf{W}^\top \mathbf{\Lambda}\mathbf{W}\Vert_{i}>\nu\\
    &\overset{(a)}{\Longleftarrow} \vert p\vert\Vert\mathbf{W}^\top \Vert_{i}\Vert\mathbf{\Lambda}\Vert_{i}\Vert\mathbf{W}\Vert_{i}> \nu\\
    &\Longleftrightarrow \vert p\vert\Vert\mathbf{W} \Vert_{i}^2\Vert\mathbf{\Lambda}\Vert_{i}>\nu,
\end{align*}
where (a) follows from the supermultiplicativity of the infimum norm.

\item {Claim 3: Any $\mathbf{\Lambda}$ that satisfies the condition 
\begin{equation}
\label{eq:claim3}
\delta+0.5+\vert m\vert\left\Vert\mathbf{\Lambda}\right\Vert_2 \left\Vert\mathbf{W}\right\Vert_2 \leq   \Vert\mathbf{\Lambda}\Vert_{i}
\end{equation}
will ensure that equation \eqref{sn1} is satisfied.} This claim follows by noting that 
\begin{align*}
    \eqref{sn1}&\overset{(a)}{\Longleftarrow} \left\Vert(\mathbf{Q}+\mathbf{\Lambda})^{-1}\right\Vert_2\left\Vert\mathbf{S}-m\mathbf{\Lambda}\mathbf{W}\right\Vert_2 \leq   1\\
    &\overset{(b)}{\Longleftrightarrow} \left\Vert\mathbf{S}-m\mathbf{\Lambda} \mathbf{W}\right\Vert_2 \leq   \Vert\mathbf{\mathbf{Q}+\Lambda}\Vert_{i}\\
    &\Longleftrightarrow \left\Vert0.5\mathbf{I}-m\mathbf{\Lambda} \mathbf{W}\right\Vert_2 \leq   \Vert\mathbf{\Lambda}-\delta\mathbf{I}\Vert_{i}\\
    &\overset{(c)}{\Longleftarrow} 0.5+\vert m\vert\left\Vert\mathbf{\Lambda}\right\Vert_2 \left\Vert\mathbf{W}\right\Vert_2 \leq   \Vert\mathbf{\Lambda}-\delta\mathbf{I}\Vert_{i}\\
    &\Longleftrightarrow 0.5+\vert m\vert\left\Vert\mathbf{\Lambda}\right\Vert_2 \left\Vert\mathbf{W}\right\Vert_2 \leq   \Vert\mathbf{\Lambda}\Vert_{i}-\delta\\
    &\Longleftrightarrow \delta+0.5+\vert m\vert\left\Vert\mathbf{\Lambda}\right\Vert_2 \left\Vert\mathbf{W}\right\Vert_2 \leq   \Vert\mathbf{\Lambda}\Vert_{i},
\end{align*}
where (a) and (c) follow from submultiplicativity and subadditivity, respectively, of the spectral norm, while (b) follows from the relation $\Vert\mathbf{A}^{-1}\Vert_2=\Vert \mathbf{A}\Vert_{i}^{-1}$.
\item {Claim 4: Any $\mathbf{\Lambda}$ that satisfies the condition
\begin{equation}\label{4prop}
0.5+\vert m\vert\left\Vert\mathbf{\Lambda}\right\Vert_2 \left\Vert\mathbf{W}\right\Vert_2 \leq    \vert p\vert\left\Vert \mathbf{W}\right\Vert_{i}^{2}\Vert\mathbf{\Lambda}\Vert_{i}-\nu
\end{equation}
will ensure that \eqref{sn2} is satisfied.} This claim follows by noting that
\begin{equation*}
\begin{aligned}
    &\eqref{sn2}\Longleftarrow  \left\Vert \left(p\mathbf{W}^\top\mathbf{\Lambda}\mathbf{W}+\mathbf{R}\right)^{-1}\right\Vert_2 \left\Vert\mathbf{S}-m\mathbf{\Lambda} \mathbf{W}\right\Vert_2 \leq   1\\
    &\Longleftrightarrow \left(\left\Vert\mathbf{S}\right\Vert_2+\vert m\vert\left\Vert\mathbf{\Lambda}\right\Vert_2 \left\Vert\mathbf{W}\right\Vert_2\right) \leq   \left\Vert p\mathbf{W}^\top\mathbf{\Lambda}\mathbf{W}+\mathbf{R}\right\Vert_{i}\\
    &\Longleftrightarrow 0.5+\vert m\vert\left\Vert\mathbf{\Lambda}\right\Vert_2 \left\Vert\mathbf{W}\right\Vert_2 \leq   \left\Vert p\mathbf{W}^\top\mathbf{\Lambda}\mathbf{W}-\nu\mathbf{I}\right\Vert_{i}\\
    &\Longleftrightarrow 0.5+\vert m\vert\left\Vert\mathbf{\Lambda}\right\Vert_2 \left\Vert\mathbf{W}\right\Vert_2 \leq   \left\Vert p\mathbf{W}^\top\mathbf{\Lambda}\mathbf{W}\right\Vert_{i}-\nu\\
    &\Longleftarrow 0.5+\vert m\vert\left\Vert\mathbf{\Lambda}\right\Vert_2 \left\Vert\mathbf{W}\right\Vert_2 \leq   \vert p\vert\left\Vert \mathbf{W}\right\Vert_{i}^{2}\Vert\mathbf{\Lambda}\Vert_{i}-\nu.
\end{aligned}
\end{equation*}
\end{enumerate}
Since~(\ref{eq:claim3}) is a sufficient condition for~(\ref{Lmin}) and~(\ref{4prop}) is a sufficient condition for~(\ref{boundclaim2}), we have shown that if there exists a matrix $\mathbf{\Lambda}$ of the form~(\ref{lambdadef}) that ensures that the two conditions conditions~(\ref{eq:claim3}) and~(\ref{4prop}) are satisfied, then the LMI~(\ref{sprocineqstate}) is feasible for that $\mathbf{\Lambda}$. The proof now follows by noting that if~(\ref{eq:thm_cond_1}),~(\ref{eq:thm_cond_2}), and~(\ref{eq:thm_cond_3}) hold, then~(\ref{eq:claim3}) and~(\ref{4prop}) are satisfied.
\end{proof}

This result provides a computationally easy to impose condition that guarantees sector boundedness of the mapping defined by each neural network layer in terms of the spectral norm of the weight matrix. The conditions for each layer can be combined to guarantee the sector boundedness of the entire neural network following Theorem~\ref{thm:entire_network}.  While this theorem provides only a sufficient condition, a few observations can be made from how easy it is to satisfy the condition~(\ref{eq:thm_cond_1}). 
\begin{itemize}
    \item If $p=0$ (as is the case with ReLU for instance), then the bound on the right hand side of this condition is smaller. This implies that considering leaky ReLU (for which $p\neq 0$) may lead to superior robustness performance as compared to ReLU activation functions, as has been observed empirically in the literature \cite{xu2015empirical}.
    \item Similarly, if $\nu>0$, then the condition becomes harder to satisfy.  Thus, imposing strict passivity on the neural network ($\nu>0$) may be overly conservative and lead to low performance as compared to simply sector bounding it and allowing $\nu<0$.
    \item We emphasize that although the result aligns with the empirical observation in the literature that regularizing the loss function with the spectral norm of the weight matrix leads to superior robustness against adversarial attacks, our result provides an analytical justification of such a procedure and further identifies the region in which the spectral norm should be bounded to get such robustness. \rev{Furthermore, by not using an SDP approach, we can expand our technique to deep Neural Network structures, which is a shortcoming of SDP methods, as discussed in \cite{pauli2022}. However, the trade-off is an increased conservatism in the spectral norm bound.}
    \item Finally, although our motivation in this paper was to ensure  robustness against adversarial perturbations, imposing passivity and sector boundedness on neural networks is of independent interest. For instance, this result can be used to guarantee stability of a system where the controller is implemented as a neural network through standard results in passivity based control.
\end{itemize}

\section{Experimental Results}\label{results}

We now present the experimental setup and performance improvement when the results above are utilized while training a neural network. Implementation is provided in Python using Tensorflow 1.15.4 on \href{https://github.com/beaquino/Robust-Design-of-Neural-Networks-using-Dissipativity}{https://github.com/beaquino/Robust-Design-of-Neural-Networks-using-Dissipativity}. A MATLAB script to obtain sector bounds for each layer to satisfy Theorem~\ref{thm:entire_network} is also available.

%
We use two commonly known data sets for image classification, MNIST \cite{deng2012mnist} and CIFAR-10 \cite{krizhevsky2009learning}. For the MNIST data set, we use a 3 layer feed-forward network with leaky ReLU activation function (with $a=0.1$, and therefore $\alpha=0.1$ and $\beta=1$) on the first two layers. For the CIFAR-10 dataset, we use an Alexnet \cite{alexnet}, which is composed of 2 convolution layers, each followed by a max pooling layer and 3 fully connected layers. Leaky ReLU activation function (with $a=0.1$, and therefore $\alpha=0.1$ and $\beta=1$) are used on the first four layers. In some implementations, a final Softmax layer may be utilized for conversion into probabilities. Since Softmax is only an exponential average, we do not consider such a layer without loss of generality. The adversarial attacks chosen for testing are the Fast Gradient Sign Method (FGSM) attack \cite{goodfellow2014explaining}, and the Projected Gradient Descent Method attack (PGDM) \cite{madry2017towards} using a range of strength $\epsilon$ in the interval $(0.1,0.5)$. Strength $0$ represents no attack.

We split training and test sets as 86\%-14\% of the dataset for MNIST and 80\%-20\% for CIFAR-10, and we train the model for $200$ epochs using Adam optimization. \rev{Parameters $\nu$ and $\delta$ should be chosen before the training procedure}. We select the pair $(\nu,\delta)=(-2,0.4)$ as the indices for the entire network, which restricts the neural network to a sector approximately between $(-0.215,0.465)$. These values were chosen as the basis of comparison, because they presented the best result among other different choices. We tuned them as hyperparameters and optimized them on the MNIST model. \rev{Given these values, we select the individual indexes $(\nu_i,\delta_i)$ for each layer, using Theorem \ref{thm:entire_network}, which are then used to calculate the spectral norm bound for each layer.}

\begin{figure}[!t]
\centering
\includegraphics[width=3.3in]{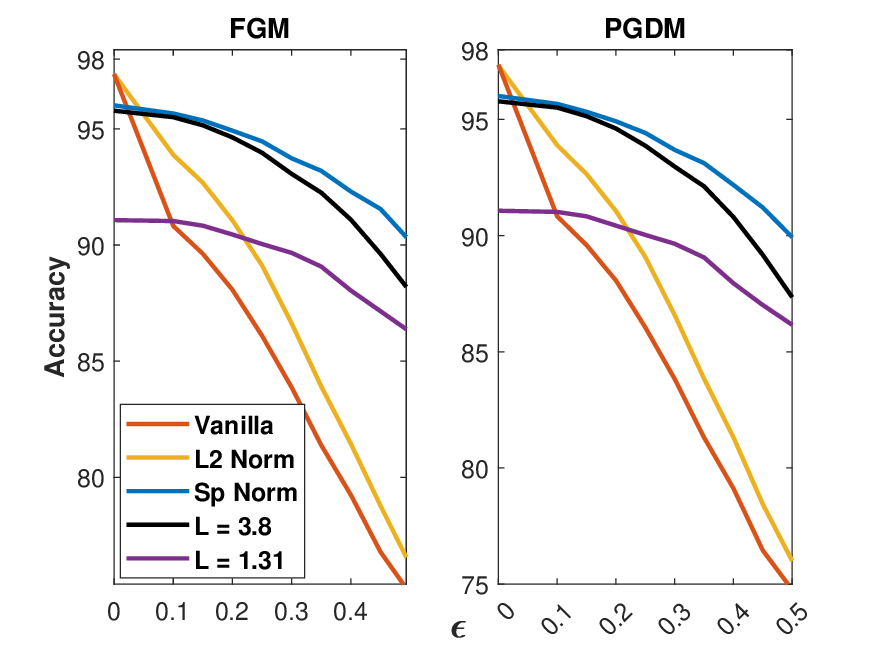}
\vspace{-5mm}
\caption{Accuracy for both Fast Gradient Sign Method attack and Projected Gradient Descent Method attack, compared for a Vanilla model and spectral norm regularized with passivity indexes $(-2,0.4)$. Network was trained on MNIST dataset. }
\vspace{-5mm}
\label{mnist}
\end{figure}
\begin{figure}[!t]
\centering
\includegraphics[width=3.3in]{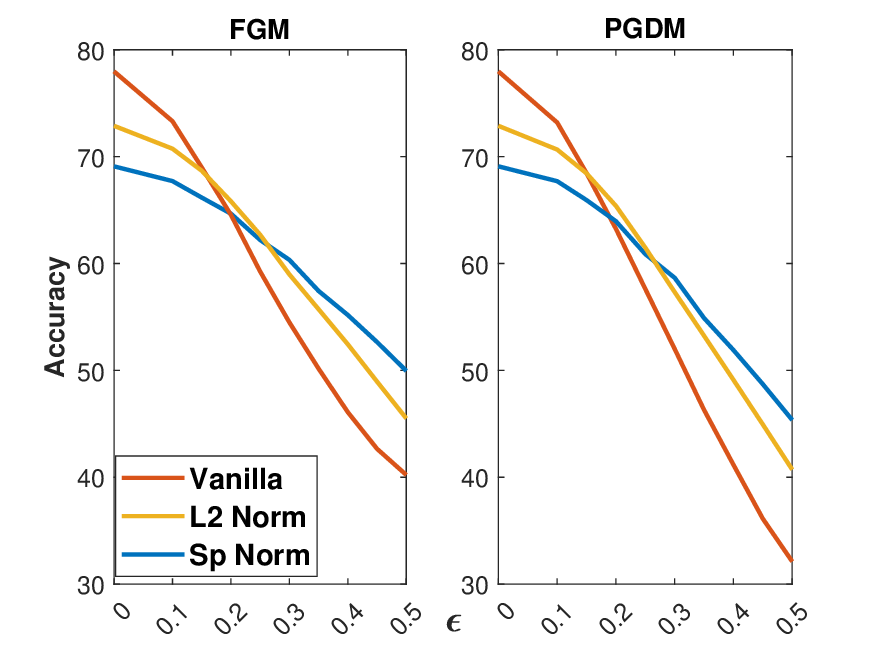}
\vspace{-5mm}
\caption{Accuracy for both Projected Gradient Descent Method attack and Projected Gradient Descent Method attack, compared for a Vanilla model and spectral norm regularized with passivity indexes $(-2,0.4)$. Network was trained on CIFAR-10 dataset.}
\vspace{-5mm}
\label{cifar}
\end{figure}

\rev{Figure \ref{mnist} presents the results (Sp Norm) for MNIST data, with the accuracy for a test set generated using FGM attack in the left panel and the accuracy for a test set generated using PGDM attack in the right.} Figure \ref{cifar} presents the results (Sp Norm) for CIFAR-10 data, with the left panel presenting the accuracy for a test set generated using FGM attack and the right panel presenting the accuracy for a test set generated using PGDM attack. Also plotted are the accuracy with the commonly used method of $L_{2}$-regularization as a comparison point (L2 Norm) and a regularly trained Neural Network (Vanilla). Both figures show an improvement on classification for both cases especially as the attack strength increases, demonstrating the effectiveness of the proposed approach.

A remark can be made about the computational tractability of the proposed approach. Considering the entire neural network at once to impose sector boundedness (or even the simpler condition of Lipschitz constant) is  computationally tractable only for shallow networks. Our approach of considering each layer separately, and crucially imposing a spectral norm constraint, is more scalable. Note that while using the spectral norm forces us to calculate the maximum eigenvalue of a matrix at every gradient descent step, this can be performed efficiently 
using the power iteration method. 

\section{Conclusion}
\label{sec:conclusion}
We proposed a robustness certificate for neural networks based on QSR-dissipativity. This method guarantees that a change in the input produces a  bounded change in the output, that is, the neural network function is incrementally sector bounded. 
We first expressed the certificate in the form of a linear matrix inequality. By using the compositional properties of dissipativity, we then decomposed the certificate into one for individual layers of the neural network. We also proposed a sufficient condition based on a spectral norm bound to offer a more computationally tractable problem for deep neural network structures. We presented the results for experiments using a 3 layer feed-forward network and an Alexnet structurure, trained with MNIST and CIFAR-10 respectively. Results showed superior performance when compared to vanilla training and $L_2$ regularization. 


%



 \section*{Acknowledgment}

The authors acknowledge comments from Julien Béguinot (Télécom Paris - Institut Polytechnique de Paris) and Léo Monbroussou (École Normale Supérieure Paris-Saclay - Institut Polytechnique de Paris) to update the stated assumptions in earlier versions of the draft.

\ifCLASSOPTIONcaptionsoff
  \newpage
\fi



%

\bibliographystyle{IEEEtran}
\bibliography{mybib}

\end{document}